\documentclass[runningheads]{llncs}
\usepackage[utf8]{inputenc}
\usepackage[T1]{fontenc}
\usepackage{graphicx}

\usepackage{mathptmx} 
\usepackage{amsmath} 
\usepackage{amssymb} 

\usepackage{bm}
\usepackage{booktabs}
\usepackage[ruled,vlined,boxed]{algorithm2e}
\usepackage{multirow}
\begin{document}
%
\title{Enhancing Counterfactual Image Generation Using Mahalanobis Distance with Distribution Preferences in Feature Space}
%
%
\author{Yukai Zhang\inst{1} \and
Ao Xu\inst{2} \and
Zihao Li\inst{3} \and
Tieru Wu \inst{4} 
}

\institute{Jilin University 
\email{ykzhang22@mails.jlu.edu.cn} \and
Jilin University 
\email{xuao22@mails.jlu.edu.cn} \and
Jilin University 
\email{zihaol20@mails.jlu.edu.cn} \and
Jilin University 
\email{wutr@jlu.edu.cn} 
}


%
\maketitle              
\begin{abstract}
In the realm of Artificial Intelligence (AI), the importance of Explainable Artificial Intelligence (XAI) is increasingly recognized, particularly as AI models become more integral to our lives. One notable single-instance XAI approach is counterfactual explanation, which aids users in comprehending a model's decisions and offers guidance on altering these decisions. Specifically in the context of image classification models, effective image counterfactual explanations can significantly enhance user understanding. This paper introduces a novel method for computing feature importance within the feature space of a black-box model. By employing information fusion techniques, our method maximizes the use of data to address feature counterfactual explanations in the feature space. Subsequently, we utilize an image generation model to transform these feature counterfactual explanations into image counterfactual explanations. Our experiments demonstrate that the counterfactual explanations generated by our method closely resemble the original images in both pixel and feature spaces. Additionally, our method outperforms established baselines, achieving impressive experimental results.

\end{abstract}

\section{Introduction}

In recent years, substantial growth and development in AI \cite{whittaker2018ai} have positioned it as a pivotal driver of progress across diverse application areas. A primary challenge in the field of machine learning, particularly with the shift towards complex technologies like ensembles and deep neural networks, is enhancing interpretability. This interpretability issue contrasts starkly with the earlier AI era, which was dominated by expert systems and rule-based models. The General Data Protection Regulation (GDPR) \cite{doi:10.1080/1097198X.2019.1569186} exemplifies this shift. It is a significant European regulation that underscores an individual's right to comprehend and question algorithmic decisions, thereby elevating interpretability to a vital concern for organizations embracing data-driven decision-making (European Union, 2016). Moreover, as reliance on opaque machine learning models in critical decision-making sectors increases, demands for AI transparency and accountability intensify \cite{castelvecchi2016can}. Providing comprehensive explanations for model outputs is crucial in key areas such as precision medicine, finance, and transportation, where professionals require detailed insights beyond mere predictive outcomes to guide their decisions. This necessity is particularly pressing in scenarios where decisions have significant societal impacts \cite{thiagarajan2022training}. The rapid evolution of XAI in recent times is demystifying black-box models, thereby enhancing user trust and engagement. XAI caters to a wide array of black-box models, offering tailored interpretability techniques to meet diverse user requirements and improve the overall transparency of AI systems \cite{arrieta2020explainable}.

\begin{figure}[t]
    \centering
    \includegraphics[width = 0.7 \linewidth ]{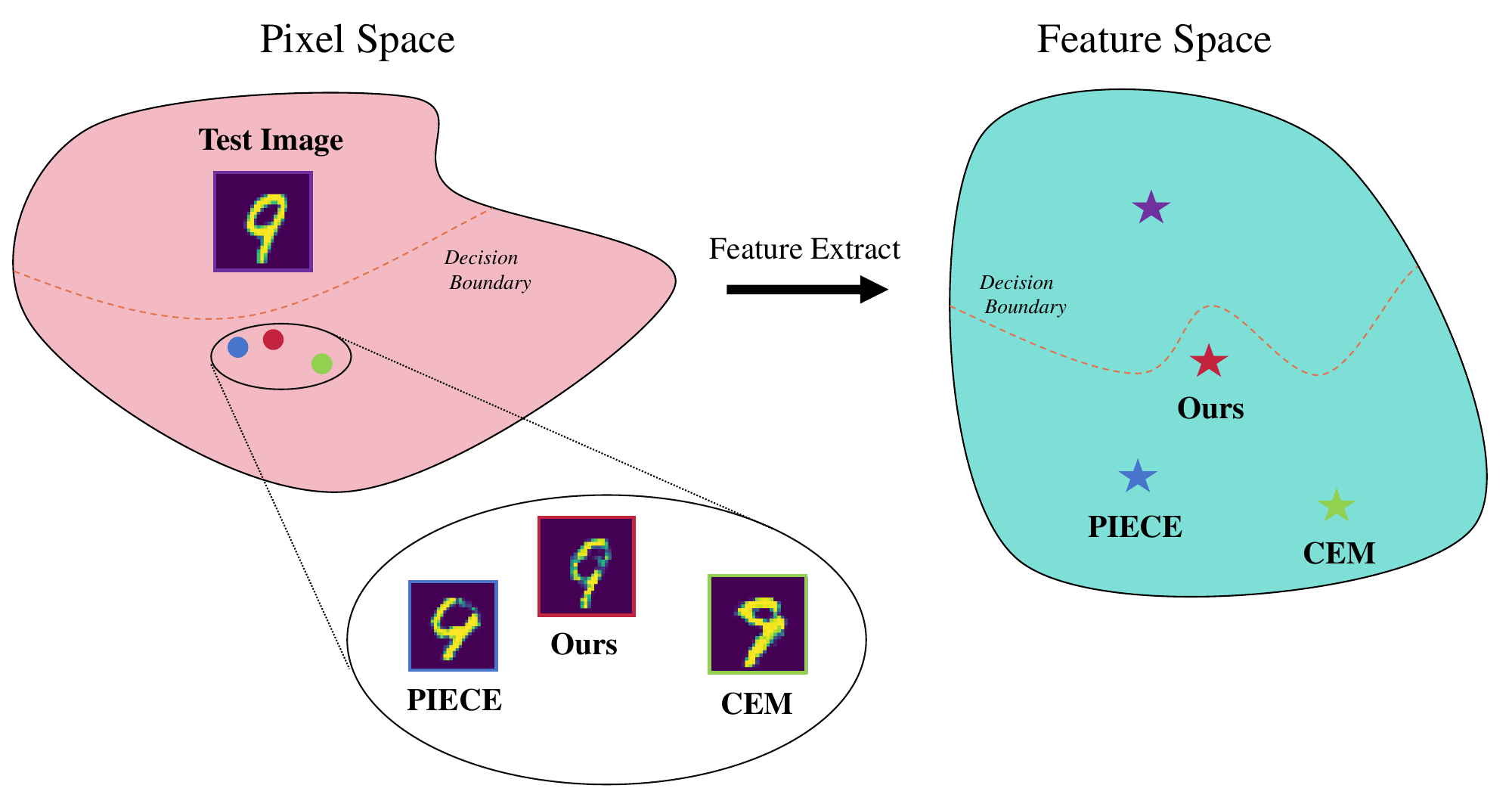}
    \caption{Our approach generates image counterfactual explanations that remain close to the original image in pixel space while also maintaining a sufficiently small distance from it in feature space.}
    \label{figintro}
\end{figure}
 
The authors of \cite{wachter2017counterfactual} introduced the concept of counterfactual explanations, which focus on identifying minimal feature changes necessary to alter a model's predictions. These explanations offer valuable insights for users, helping them comprehend the decision-making processes of black-box models. Previous research, as documented in \cite{pawelczyk2020learning}, \cite{van2021interpretable}, and \cite{guyomard2022vcnet}, has presented a variety of algorithms for generating such explanations. Specifically, a counterfactual explanation guides the user on changes required for an input to shift the model's prediction from class $C_1$ to class $C_2$, incurring minimal cost. An easy to understand example of the role of counterfactual explanation is as follows:
\begin{quote}
    \textit{Your loan application was denied; however, if you increase your credit score by $2$ points, your application will be approved.}
\end{quote}
From an algorithmic perspective, generating counterfactual explanations requires identifying an output instance that meets three key conditions \cite{hamman2023robust}:

\begin{itemize}
\item Proximity to the input instance.
\item Minimization of feature changes.
\item Adherence to the underlying data distribution.
\end{itemize}

Counterfactual explanations are primarily applied to tabular and image data. Tabular data, usually containing partially discrete features, is handled through encoding techniques. In counterfactual explanations, providing a feature vector of the same dimension as the input instance suffices. For image data, the main goal is to identify and modify abnormal features in the input image to resemble a normal image, as discussed in \cite{kenny2021generating}. This is consistent with the general understanding of counterfactual explanations.

\subsection{Our Contributions}


In this paper, we make the following contributions. 

\begin{itemize}
    \item Propose that counterfactual explanations of image datasets should be found in the feature space of black-box models rather than the pixel space.
    \item Propose an algorithm to compute the feature importance of all features in the feature space of the black-box model.
    \item Use the distance function which combines the Mahalanobis distance and the feature importance to compute the counterfactual explanation in the feature space, and generate the image counterfactual explanation by the generator.
\end{itemize}

\subsection{Related Work}
\subsubsection{Mahalanobis Distance.}
The Mahalanobis distance measures the distance between a point $P$ and a distribution $D$ and is commonly used for outlier detection. Previous work, such as \cite{wachter2017counterfactual}, has treated the resolution of counterfactual explanations as an optimization problem incorporating distance constraints between input and counterfactual instances. However, the authors of \cite{kanamori2020dace} suggest employing the Mahalanobis distance along with the outlier detection method LOF \cite{breunig2000lof} to constrain the distance between counterfactual and input instances.

\subsubsection{Image Counterfactual Explanation.}
For generating counterfactual explanations in image datasets, it is necessary to provide image counterfactual explanations instead of feature vectors or other types of explanations. Generally, these image counterfactuals are derived in feature space and subsequently transformed into images via an image generator for user comprehension. In previous research, CEM \cite{dhurandhar2018explanations} seeks counterfactual explanations directly in pixel space and utilizes an AutoEncoder (AE) to ensure generated images align with the original dataset's distribution. Building on CEM, Proto-CF \cite{van2021interpretable} proposes using class centers in the AutoEncoder's feature space of the target class to expedite counterfactual explanation generation. PIECE \cite{kenny2021generating} creates counterfactual explanations in feature space, generating new features based on the input instances' proximity to the target category's distribution in the feature space, and then maps these features back to image space using a GAN's generator \cite{goodfellow2014generative}. However, while DVCEs \cite{augustin2022diffusion} produce superior counterfactual images, they overlook the primary goal of interpretability, which is to explain black-box models. DVCEs focus solely on generative tasks, without considering black-box models.

\section{Preliminaries}
\subsection{Wasserstein Distance}
In our notation, we denote $\mathcal{X}$ as the instance space, and consider a metric:
\begin{definition}
    Let $d(\cdot,\cdot):\mathcal{X}^2\rightarrow \mathbb{R}_+$ be a metric. A space $(\mathcal{X},d)$ is Polish if  it is complete and separable. Throughout it is assumed that all Polish spaces $(\mathcal{X},d)$ are equipped with the Borel $\sigma$-algebra $\mathcal{E}$ generated by $d(\cdot,\cdot)$.
\end{definition}
Based on this metric, we can establish the definition of the Wasserstein distance \cite{villani2009optimal}:
\begin{definition}
    Let $(\mathcal{X},d)$ be a Polish metric space and let $p\in [1,+\infty)$.  Then, the Wasserstein distance of order $p$ between two probability distributions $P$ and $Q$ on $\mathcal{X}$ is
    \begin{align}
\mathbb{W}_p(P,Q)=\left ( \inf_{R\in \Pi(P,Q) }\int _{\mathcal{X}^2  }d(x,y)^pdR(x,y) \right )^{1/p} ,
\end{align}
where $\Pi(P,Q)$  is the set of all couplings $R$ of $P$ and $Q$, i.e., all joint distributions on $\mathcal{X}^2$ with marginals $P$ and $Q$. 
\end{definition}

\begin{figure*}[ht]
    \centering
    \includegraphics[width = \linewidth]{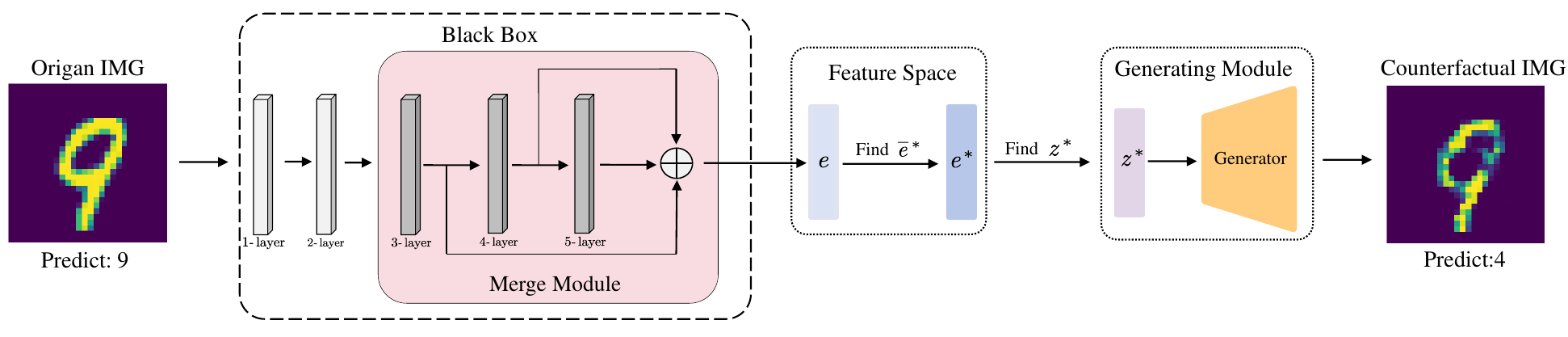}
    \caption{For the black box and the Origan IMG predicted to be 9, we first select a number of layers in the black box that require feature fusion, perform fusion of input instance features and fusion of distribution information vectors, then use the two strategies we devised to find the most appropriate counterfactual explanation category, then solve for the optimal counterfactual explanations in the feature space, and finally find the optimal inputs for a generative module to obtain the Counterfactual IMG.}
    \label{fig:enter-label}
\end{figure*}

The Wasserstein distance is a mathematical metric used to quantify the similarity or dissimilarity between two probability distributions. It measures the minimum average cost required to transport one distribution to another, with the cost defined based on the distance between the two distributions. The concept of Wasserstein distance originates from the fields of transportation problems and convex optimization. Its significance in applications lies in providing an effective means to quantify differences between various distributions while possessing desirable properties such as submodularity and the triangle inequality, making it a valuable tool in the field of machine learning. It aids in addressing numerous practical problems, including data matching, image synthesis, domain adaptation, and more.
\subsection{Counterfactual Explanation}
The counterfactual explanation concerning instance $x$ and model $f$ can be defined as the minimal perturbation to $x$ resulting in a change to the prediction, denoted as $\Delta_x$. 
we define the optimal counterfactual example as
\begin{align*}
\bar x^*:= \underset{\bar x }{\operatorname{argmin }} \    d(x,\bar x)\text{ s.t. }f(x)\neq f(\bar x).
\end{align*}
The corresponding optimal counterfactual explanation is $\Delta_x^*=\bar x^*-x$. This definition is consistent with prior research in machine learning regarding counterfactual explanations \cite{laugel2019issues}. Meanwhile, Wachter, Mittelstadt, and Russell \cite{wachter2017counterfactual} adopt an alternative perspective, utilizing gradient descent to seek counterfactual explanations. Specifically, they have devised the following loss function:
\begin{align}\label{eqdefCE}
\mathcal{L}(x,\bar{x}|f,d ) & = \mathcal{L}_{pred}(x,\bar{x}|f)  +\lambda\cdot \mathcal{L}_{dist} (x,\bar{x}|d).
\end{align}
The loss function consists of two terms with a weight $\lambda>0$. The first term represents the predictive loss, which forces the counterfactual instance $\bar{x}$  to have a different prediction category than the input instance $x$. The second term represents the distance between input instance and counterfactual instance, with a smaller value encouraging $\bar{x}$ to move closer to the decision boundary.
The underlying premise is that an optimal counterfactual example, denoted as $x^*$, can be determined through the minimization of the comprehensive loss function.

\section{Main Method}

\subsection{ Notations and Overview}
\label{secover}
Assuming that we consider a training dataset of size $M$, denoted as $D=\{(x_i, y_i)\}_{i=1}^M$. Consider a classification model, denoted as $f=E \circ \sigma\circ A$, comprised of $D+1$ layers, where
$$
E=W_1\circ \sigma \circ W_2\circ \sigma\circ \cdots \circ W_{D-1}\circ \sigma \circ W_D,
$$
where $W_i$ represents the $i$-th layer of $E$, $\sigma$ represents the activation function, and $A$ is a linear layer that we employ to obtain the final output. In this context, we assume that the dimensions of the outputs for each layer of $E$ are denoted as $2^{N_1}, \cdots ,2^{N_D}$, and the output of the $j$-th neuron in the $i$-th layer is represented as $W_{i,j}$, where $i=1, \cdots, D$ and $j=1, \cdots, 2^{N_i}$. For the sake of notational simplicity, we also use $W_{i,j}$ to refer to the corresponding neuron and $W_{i,j}(x)$ to denote the output of that neuron after input instance $x$. In addition, let us consider our task as a multi-class classification problem with labels $C_1, \cdots, C_K$. 

The primary methodology proposed in this paper can be divided into two main components. Firstly, we will introduce the first part in Section \ref{secfeature}, which deals with the selection and fusion of feature layers. We employ the Wasserstein Distance to define the Passing Rate metric, which aids in identifying the crucial feature layers. Then, we merge the selected feature layers using three distinct merging strategies: balanced merging, weakening merging, and strengthening merging.
Secondly, in Section \ref{secoptim}, we introduce the concept of Distribution Preference Mahalanobis Distance and employ it to devise a novel loss function for solving counterfactual explanations.
\subsection{The Selection and Fusion of Feature Layers}
\begin{figure*}[h]
    \centering
    \includegraphics[width=\linewidth]{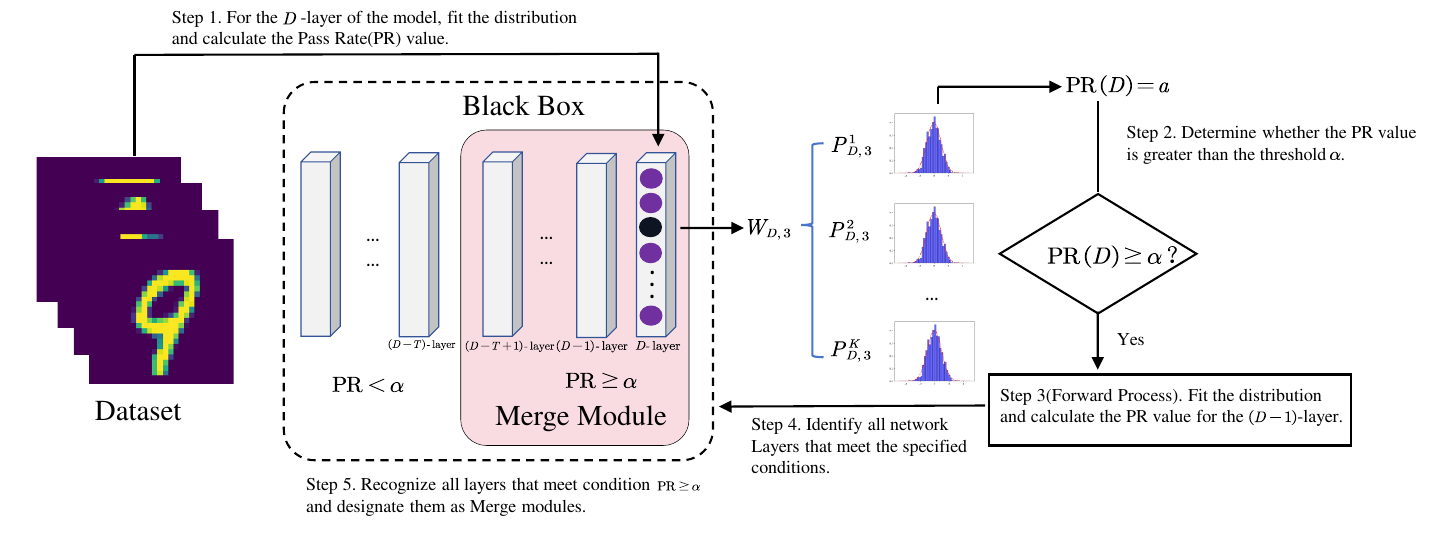}
    \caption{The process of finding Merge Modlue.}
    \label{fig:feature}
\end{figure*}
\label{secfeature}
We divide the training dataset into $K$ subsets based on the predictions of the data therein under the black-box model: 
$$D_l=\{(x, y)\in D: f(x)=C_l\}, $$
where $l=1, 2, \cdots, K$. For each $D_l$, we can obtain |$D_l|$ outputs for the neuron $W_{i,j}$. 
By utilizing these $|D_l|$ outputs, we can employ a fitting method to obtain a distribution associated with the neuron $W_{i,j}$, denoted as $\mathcal{P}_{i,j}^l$. Consequently, each $D_l$ corresponds to $D$ vectors:
$$
\big(\mathcal{P} _{1,1}^l, \mathcal{P}_{1,2}^l,\ldots, \mathcal{P}_{1,2^{N_1}}^l\big), \ldots, \big(\mathcal{P}_{D,1}^l, \mathcal{P}_{D,2}^l, \ldots, \mathcal{P}_{D,2^{N_D}}^l\big).
$$

We utilize this distribution information to select important feature layers, and this selection primarily relies on the meaningfulness of each layer in a semantic context.
In a specific context, we define the following metric as the Passing Rate:
\begin{align}\label{eqpr}
\text{PR}(e) = \frac{1}{K \cdot 2^{N_e}}  \sum_{j = 1}^{2^{N_e}} \sum_{l = 1}^{K}  \text{PV}\big(D_l, \mathcal{P}^l_{e,j}\big),
\end{align}
where $\text{PV}\big(D_l, \mathcal{P}^l_{e,j}\big)$ represents the p-value of $W_{i,j}(x)$ with respect to the distribution $\mathcal{P}^l_{e,j}$, and $e$ is the index of the layer.
Intuitively, this metric assesses the reasonableness of distribution fitting operations for layer $e$. Using this metric, we calculate the Passing Rate from the last layer in a backward manner until the Passing Rate value is less than confidence $\alpha$. Let $\#$ be the number of layers retained after this filtering process. The final number of selected feature layers, calculated from the back to the front, is given by:
$$ T=\max\{\min\{\#, \text{max}\_\text{num}\}, 1\},$$
where $\text{max}\_\text{num}$ is a constant, and in this paper, it is taken as $\left \lfloor D/2 \right \rfloor$. This approach ensures that the final number of selected layers is neither excessive nor too limited. Fig. \ref{fig:feature} briefly expresses how to select and fuse feature layers.

Following the above operations, we obtain the selected feature layers $\mathcal{W} = \{W_{D-T+1},\cdots, W_D\}$.
Then, we merge these layers in three ways. First, we choose $V$ as 
$\min\{2^{N_{D-T+1}}, \cdots, 2^{N_D}\},$
and then use average pooling to pool each vector in $\mathcal{W}$ into a feature of length $V$, denoted as
$ \mathcal{W}' = \{W_{D-T+1}',\cdots, W_D'\}$.
Finally, we merge the elements in $\mathcal{W}'$.
We consider three different merging strategies, namely, forward increasing weights $(a > 1)$, equal weights $(a = 1)$, and forward decreasing weights strategies $(a < 1)$.
$$W_{Feature}=\frac{1}{T}\sum_{i=0}^{T-1}a^i\cdot W'_{D-i}.$$


\subsection{Using DPMD to Resolve Counterfactual Feature Attribution}
\label{secoptim}
The formulation of the loss function component $\mathcal{L}_{dist}$ in Equation (\ref{eqdefCE}) holds significant importance. Existing literature has extensively explored loss functions that depend on various metrics. Kentaro Kanamori et al. \cite{kanamori2020dace} introduced a loss function based on the Mahalanobis distance, which is represented as
\begin{align}
\mathcal{L}_{dist} (x,\bar{x}|d)=d_{M}(x,\bar{x})=\sqrt{(x-\bar{x})^{\top }\Sigma \left (  x-\bar{x}\right )  }
\label{eqMdist}
\end{align}

where $\Sigma\in \mathbb{R}^{V\times V}$ is a semi-definite matrix and achieved favorable experimental results. However, their loss function does not take into account the importance of each coordinate in the instance space. We utilize distribution information to characterize this importance and improve Equation (\ref{eqMdist}). 

\begin{figure}[h]
    \centering
    \includegraphics[width  = 0.8 \linewidth]{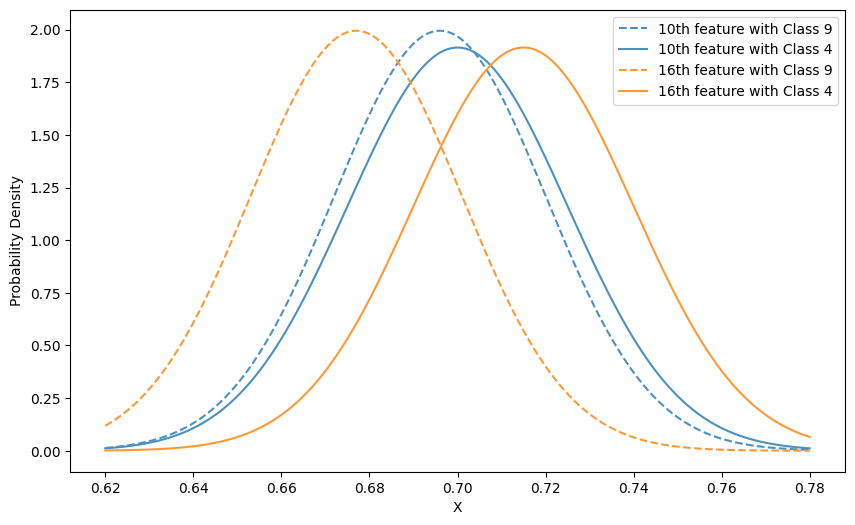}
    \caption{Distribution of data from data class 9 and data from data class 4 on the 10th neuron and 16th neuron, respectively.}
    \label{fig:featureimportance}
\end{figure}
Fig. \ref{fig:featureimportance} depicts four distribution curves obtained during our experimental process, corresponding to the 10th and 16th features of categories 9 and 4, respectively. Assuming our task is to find an output instance belonging to category 4 for an input instance belonging to category 9, we observe that the distribution distance of the 10th feature between categories 9 and 4 is small. This suggests that changing the value of the 10th feature in the input instance makes it challenging to transition from category 9 to 4. In contrast, the distribution distance of the 16th feature between categories 0 and 4 is large. Specifically, category 4 typically exhibits larger values for the 16th feature, while category 9 usually has smaller values for the 16th feature. Therefore, increasing the value of the 16th feature is likely to significantly alter the category of the input instance. We define this distribution distance as the feature importance between categories 9 and 4, where features with greater importance are assigned higher weights during optimization.

Suppose we aim to find the counterfactual explanation for an input instance with label $l$ regarding label $l'$. For any element $W_i$ in $W = \{W_{D-T+1},\cdots, W_D\}$, we can compute a corresponding vector of the same dimension, denoted as $\Lambda_i$, where the $j$-th element of $\Lambda_i$ is $\mathbb{W}(\mathcal{P}^l_{ij}, \mathcal{P}^{l'}_{ij})$. Thus, we obtain a collection of vectors:
$$\tilde{\Lambda} = \{ \Lambda_i |i = D - T + 1, \cdots, D \} .$$
Through the integration method described in Section \ref{secfeature}, we obtain a final feature importance vector:
\begin{align*}
\Lambda_{feature}=(\lambda_1,\lambda_2,\cdots,\lambda_V). 
\end{align*}
Now, we enhance Equation (\ref{eqMdist}) using $D_{feature}$ as follows:
\begin{align}
\mathcal{L}_{dist} (x,\bar{x}|d) & =d_{DPM}(x,\bar{x}):= \sqrt{(x-\bar{x})^{\top }\Sigma^{'}
 \left (  x-\bar{x}\right )  }
\label{eqMdist2}
\end{align}
where $\Sigma^{'}=\Sigma+\beta\cdot \mathrm{diag}(\lambda_1,\lambda_2,\cdots,\lambda_V )$, $\beta$ is a balance parameter. We denote $d_{DPM}(\cdot,\cdot)$ as the Distribution Preference Mahalanobis Distance.

\subsection{How to find the label of optimal counterfactual}
\label{secBestMatch}
Typically, the user needs to specify the class of counterfactuals to be generated, but we want our algorithm to be able to generate the most appropriate counterfactuals even if they are not specified by the user. We propose two strategies for finding the most appropriate category of counterfactual explanations in our framework.  \par

\textbf{Strategy 1.} Assuming that the predictive label of the input instance is $C_1$, we determine the category of counterfactual explanation by comparing the similarity of the other categories to the category $C_1$ in the feature space, which can be defined as the sum of the distributional distances corresponding to all neurons of the two categories in the feature space. Let the feature dimension be $n$. 
The strategy can be expressed as
\begin{align}
    j^* = \underset{j\in\{1,2,\ldots,K\}}{\text{argmin }}  \sum_{i=1}^{2^D}\mathbb{W}\big(\mathcal{P}^{C_1}_{D,i}, \mathcal{P}^{C_j}_{D,i}\big),
\end{align}
where $j^*$ is the most appropriate category of counterfactual explanations under this strategy. \par

\textbf{Strategy 2 (Proto-Class).}  Suppose $E$ is a feature extractor for a black-box model as a mapping from pixel space to feature space. For each class $j$ that is not $C_1$, we map all real instances belonging to this class to the feature space, and then sort them in ascending order of $L_2$ distance from $E(x_0)$ in the feature space, where $E(x_0)$ is the feature of the input instance in the feature space, then the optimal counterfactual explanation class sought by this strategy can be defined as
\begin{align}
    j^* = \underset{j\neq C_1}{\text{argmin }} \| E(x_0) - \text{proto}_j\|^2_2,
\end{align}
where $\text{proto}_j = \frac{1}{K}\sum_{k=1}^K E(x_k^j)$ is the class center of the top-K nearest instances of $E(x_0)$ belonging to class $j$.

By using the above strategy, we are able to solve for the optimal category of counterfactual explanation $j^*$, thus guiding and accelerating the generation of counterfactual explanation.

\subsection{Using DPMD to Generate Image Counterfactual Explanation}
\begin{algorithm}[h]
\label{alrorithm1}
\caption{Data Process}
\KwIn{The black box model $f$. The Train Dataset $D$. Number of classes $K$. The Threshold value of forward process $\alpha$. The $j$-ths layer of DNN, and its number of neurons $(W_j,V_j),$ $j=1,2,...,D$}

\nlset{1} \textbf{Save the data according to the labels predicted by the black box model.}\\
\ForEach{data in $D$}{
compute label $l =f(data)$\\
append $W_D(data)$ to $DF^l$ 
}

\nlset{2}\textbf{Fitting Distribution In Lantent Space.}\\
$Count = 0$ \\
\For{$l \leftarrow 1 $ to $ K$}{\For{$i \leftarrow 1 $ to $ V_D$}{$FitData = DF^l_i$ \\
Fitting\\
Save \textit{ks-p-value, loc, lic} etc. \\
\If{ks-p-value > 0.05}{
   $Count += 1$ 
}}}
compute $PassRate = Count/(K*V_D)$ \\
\nlset{3}\textbf{Forward Process.} \\
\While{$ PassRate \geq \alpha$ }{
1-2 with $(W_j,V_j), j = D-1,D-2,...,1$
}


\end{algorithm}

We have determined the class of counterfactual explanations and the distance function, then, in the feature space $\mathcal{E}$ , we can apply the following equation to solve for the counterfactual explanations:

\begin{align} \label{eq8}
\bar e^*:=  \underset{\bar e\in \mathcal{E} }{ \text{argmin}} \    \mathcal{L}_{dist} (e,\bar{e}|d_{DMP})     \text{ s.t. }A( \bar e^* )=j^*,
\end{align}
where $e$ is the feature of input instance, $j^*$ is the optimal counterfactual class derived in Section \ref{secBestMatch}.


After deriving the feature counterfactual explanation in feature space, our next objective is how to map the feature counterfactual explanation to pixel space. To solve this problem, similar to the work of PIECE, we use the generator GAN, whose input is a noise vector $z$. The generated image is required to meet the following conditions: (a) exhibiting minimal deviation in feature space from the feature vectors obtained through Equation (\ref{eq8}), (b) displaying minimal differences in pixel space, and (c) retaining the predefined labels. The solution equation is as follows:
\begin{align} \label{eq9}
    \begin{split}
    z^* = \underset{z}{\text{argmin }} &\Big\{|| E(G(z)) - \bar e^* ||_2^2 + \mu  ||G(z) - x||_2^2 \Big\} \\
    \text{ s.t. } & f(G(z))=j^*.
    \end{split}
\end{align}
 When $z^*$ is solved by the above equation, our image counterfactual explanation can be expressed as $I^*=G(z^*)$.


\section{Experiments}

We conduct experiments on real datasets to investigate the effectiveness of our DPMDCE by comparing the performance with existing methods for generating counterfactual explanation. 
The dataset used for our experiments is the MNIST dataset\cite{lecun-mnisthandwrittendigit-2010}, whose training set contains 60,000 images and the test set contains 10,000 images, each of which is a single-channel pixel size of 28*28, and whose labels are integers between 0 and 9.




\subsection{Baselines} \label{secBaselines}
\paragraph{\textbf{Min-Edit.}} The minimal editing method for image counterfactual explanation
is to let the pixels be perturbed directly in the pixel space of the input instance until the label of the image is changed.
It can be written as the following equation:
\begin{align*}
    x^* = \underset{x^\prime}{\text{argmin }} d(x,x')
    \text{ s.t. } f(x)\neq f(x^\prime).
\end{align*}

\par

\paragraph{\textbf{PIECE.}}
PIECE first obtains the features of all the training set data at the feature layer, then fits all the features belonging to the same category into a distribution according to the neuron in which they are located, and then determines in turn whether each feature of the input instance belongs to an outlier (greater than the upper t-quantile or less than the lower t-quantile) in the distribution of the corresponding feature in the target category, and if it is an outlier assigns the value of that feature to the mean of the corresponding feature's distribution. Counterfactual explanations are then generated in the feature space and then optimization is used to convert the feature explanations into image explanations.\par
\paragraph{\textbf{CEM.}}
CEM proposes to introduce a self-encoder AE based on directly changing the pixels of the input instance so that the perturbed image is within the distribution of the original image with the following loss function:
\begin{align*}
    \delta^*   = \underset{\delta}{\text{argmin }} &\Big\{\beta \|\delta\|_1 + \|\delta\|_2^2 + \gamma \|x_0 + \delta - \text{AE}(x_0 + \delta)\|_2^2 \Big\}\\
    \text{ s.t. } &  f(x_0) \neq f(x_0 + \delta)
\end{align*}

\paragraph{\textbf{Proto-CF.}} Proto-CF adds the module of automatic selection of counterfactual explanation target categories to CEM to ensure that the algorithm can have a faster convergence rate. Its loss function is as follows:
\begin{align*}
\delta^* = \underset{\delta}{\text{argmin }} &\Big\{ \beta \cdot \|\delta\|_1 + \theta \cdot \|\text{ENC} (x_0 + \delta) - 
         \text{proto}_j||_2^2
    \\     &+\|\delta\|_2^2 + \gamma \cdot \|x_0 + \delta - 
         \text{AE}(x_0 + \delta)||_2^2\Big \}
    \\ \text{ s.t.  } & f(x_0) \neq f(x_0 + \delta)
\end{align*}
where ENC denotes the Encoder part of the AE and $\text{proto}_j$ is the class center of the identified target category in the encoder space. \par

\subsection{Evaluation Metrics}
The metrics for image counterfactual explanation should also be similar to those for tabular data, using the size of the perturbation between the original and counterfactual features, as well as the proximity to the distribution of the dataset. The variability of the image counterfactual from the original image also needs to be taken into account. 


\begin{itemize}
    \item \textbf{Fe-Dist.} The counterfactual explains the difference between the image and the original image in the latent space (lower is better).
    \begin{align*}
        \textbf{Fe-Dist} = \Vert E(G(z^*)) - E(x)\Vert_2^2.
    \end{align*}
    \item \textbf{Pixel.} The counterfactual explains the difference between the image and the original image in terms of image pixels (lower is better).
    \begin{align*}
        \textbf{Pixel-Dist} = \Vert G(z^*) - x\Vert_2^2.
    \end{align*}
    \item \textbf{Opt-Time.} The time of code run (lower is better). 
    \item \textbf{Suc-Rate.} The success rate of get one counterfactual instance which has counter label (higher is better).
    
\end{itemize}

\begin{figure*}[h]
\centering
    \includegraphics[width = \linewidth]{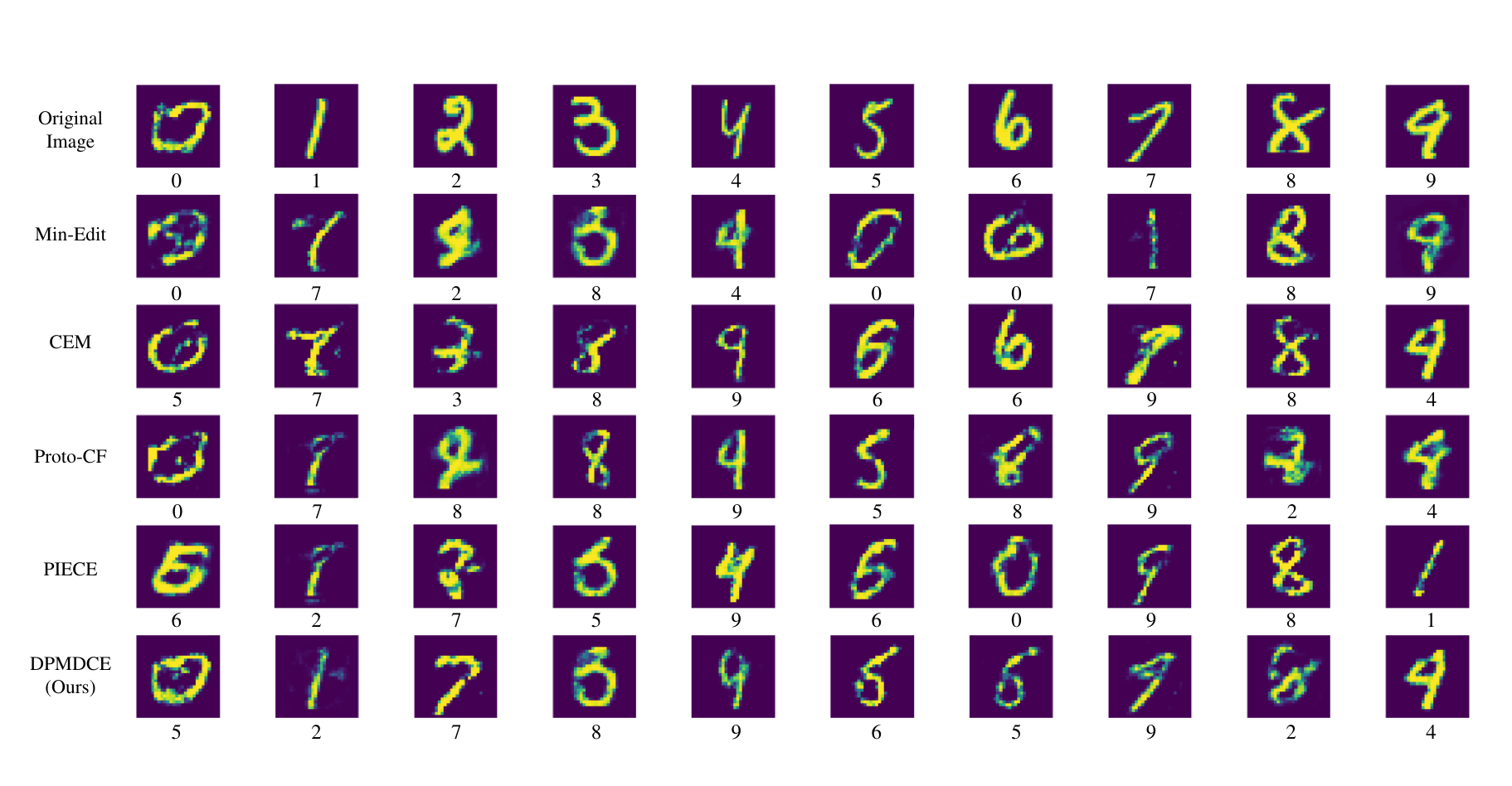}
    \caption{The first row is a randomly selected image from the MNIST test set, and the other rows are counterfactual explanations of the images generated by different counterfactual explanation generating algorithms, with the name of the algorithms labeled in the first column, and the predictions of the generated images in the corresponding black-box model labeled directly below each image. }
    \label{expfig1}
\end{figure*}

\subsection{Blackbox setup}
We use three black-box models trained on the training set of MNIST dataset, which are composed of $5$-layer fully-connected layer with ReLU activation function (Blackbox 5), 7-layer fully-connected layer with ReLU activation function (Blackbox 7), and 9-layer fully-connected layer with ReLU activation function (Blackbox 9), and their feature spaces are all of dimension 64. We use a blackbox model with accuracy greater than $70\%$. Similarly, we use the same training set to train a GAN whose inputs are 64-dimensional noise vectors.

\subsection{Experiment 1. Compared with Baselines}
We perform comparison tests with baselines on three structurally different black-box models using distance constraints with different numbers of paradigms, and we set the total number of optimization iterations for each method to 8000.\par


Fig. \ref{expfig1} shows the results of our experiments on Blackbox 7. We observe that  both our method and the baselines method produce counterfactual explanations that are able to approach the original image in pixel space, but the Min-Edit method has a high probability of generating an out-of-distribution image. From Table \ref{exp1table}, we can observe that the DPMDCE method always has the smallest metric \textbf{Fe-Dist} when metric \textbf{Pixel-Dist} is similar to the results of the other baselines methods, which is exactly where the strength of our method lies in the fact that there is no significant difference in the algorithm's running time when using the same total number of iterations, and that Min-Edit benefits from its simple loss function by having a smaller running time \textbf{Opt-Time}. From metric \textbf{Suc-Rate}, the methods using strong constraints all have higher success rates in generating counterfactual explanations, and PIECE does not achieve good results due to its weak constraints on label changes.

\begin{table*}[h]
    \renewcommand{\arraystretch}{1.3}
        \caption{Metrics comparison of five counterfactual generation algorithms on the MNIST dataset. We use the $L_1$ and $L_2$ paradigms to measure the performance of the five methods on different black-box models, respectively, with the $L_k(k=1,2)$ paradigm representing the optimization process for solving counterfactual explanations and the metrics using the $L_k$ paradigm. Each row of the table shows the performance of different algorithms on different black-box models, and each column represents the experimental metrics results of different counterfactual explanation algorithms under the same metrics. The optimal metrics for one set of comparative tests will be highlighted in bold.}
    \centering
    \scalebox{0.76}{
    \begin{tabular}{ccccccccccc} 
    \hline
    & & \multicolumn{4}{c}{$L_1$ baesd} & &\multicolumn{4}{c}{$L_2$ based}\\
    \cline{3-6} \cline{8-11}
       & Method  & \textbf{Fe-Dist} & \textbf{Pixel-Dist}  & \textbf{Opt-Time} &  \textbf{Suc-Rate} & & \textbf{Fe-Dist} & \textbf{Pixel-Dist}  & \textbf{Opt-Time} &  \textbf{Suc-Rate} \\
    \hline
     \multirow{5}{*}{ \rotatebox{90}{BlackBox 1}}  
        & Min-Edit 
        & $9.59\pm2.11$ & $12.20\pm1.63$ &\bm{$12.59\pm0.68$}  & $0.52$&
        & $10.70\pm2.39$ & $13.02\pm1.46$ & $15.56\pm0.55$ & $0.51$\\
        & CEM 
        & $7.54\pm1.60$ & $7.18\pm2.08$ & $14.98\pm1.15$ & $0.75$&
        & $6.64\pm1.94$ & $6.62\pm1.86$ & $16.67\pm0.85$ & $0.72$\\
        & Proto-CF 
        & $7.85\pm1.31$ & $7.48\pm1.23$ & $13.66\pm0.77$ & $0.86$&
        & $8.22\pm1.36$ & $6.76\pm2.20$ & \bm{$15.30\pm0.74$} & $0.76$\\
        & PIECE 
        & $6.36\pm1.39$ & $8.91\pm1.73$ & $33.33\pm1.01$ & $0.60$ & 
        & $6.42\pm1.01$ & $8.98\pm1.51$ & $33.69\pm0.46$ & $0.43$\\
        & DPMDCE(ours) 
     & \bm{$4.48\pm1.07$}  & \bm{$6.91\pm1.30$} & {$28.08\pm0.98$} & \bm{$0.92$} &
     & \bm{$3.86\pm0.76$}  & \bm{$6.75\pm1.24$} & $28.26\pm0.54$ & \bm{$0.94$}\\

    \hline
         \multirow{5}{*}{\rotatebox{90}{BlackBox 2}}  
         & Min-Edit 
        & $13.05\pm2.66$ & $9.28\pm1.04$ & \bm{$13.63\pm0.72$} & $0.41$&
        & $13.13\pm2.31$ & $10.11\pm1.83$ & \bm{$13.72\pm0.84$} & $0.53$\\
        & CEM 
        & $10.08\pm2.71$ & $6.85\pm1.92$ & $17.60\pm0.58$ & $0.72$&
        & $11.26\pm2.43$ & $8.95\pm1.60$ & $16.89\pm0.76$ & $0.82$\\
        & Proto-CF 
        & $7.94\pm1.79$ & \bm{$6.52\pm1.64$} & $14.63\pm0.77$ & $0.84$&
        & $6.20\pm1.20$ & $8.52\pm1.92$ & $14.61\pm0.81$ & $0.83$\\
        & PIECE 
        & $8.00\pm1.54$ & $9.35\pm1.59$ & $33.49\pm0.48$ & $0.49$& 
        & $7.04\pm1.43$ & $7.38\pm1.59$ & $33.75\pm0.27$ & $0.43$\\
         & DPMDCE(ours) 
         & \bm{$5.97\pm0.89$} & $7.08\pm1.79$ & $22.47\pm0.43$ & \bm{$0.91$}& 
         & \bm{$5.11\pm1.08$} & \bm{$6.97\pm1.90$} & $28.44\pm0.43$ & \bm{$0.90$}\\

    \hline
         \multirow{5}{*}{\rotatebox{90}{BlackBox 3}}  
         & Min-Edit 
        & $13.87\pm2.15$ & $12.19\pm2.44$ & \bm{$13.80\pm0.88$} & $0.50$ & 
        & $12.71\pm2.16$ & $13.50\pm2.54$ & \bm{$14.16\pm1.04$} & $0.53$\\
        & CEM 
        & $9.69\pm1.71$ & {$9.19\pm1.53$} & $17.62\pm0.78$ & $0.81$& 
        & $10.04\pm2.05$ & $8.56\pm1.38$ & $18.53\pm0.65$ & $0.87$\\
        & Proto-CF 
        & $7.67\pm1.40$ & $9.49\pm1.54$ & $15.60\pm0.85$ & $0.87$&  
        & $7.83\pm1.46$ & $7.43\pm2.40$ & $16.63\pm0.75$ & $0.84$\\
         & PIECE 
        & $8.26\pm1.54$ & $9.57\pm1.67$ & $32.69\pm1.32$ & $0.37$ & 
        &$8.06\pm 1.80$ & $9.25\pm1.35$ & $34.30\pm3.10$ & $0.43$\\
         & DPMDCE(ours) 
         & \bm{$6.27\pm 1.02$} & \bm{$8.77\pm2.07$} & $21.65\pm1.15$ & \bm{$0.90$} & 
         & \bm{$5.86\pm 1.18$} & $7.34\pm2.18$ & $34.71\pm3.12$ & \bm{$0.91$}\\
       
    \hline
    \end{tabular}
    }
    
    \label{exp1table}
\end{table*}

\subsection{Experiment 2. Reasonable verify}

The Kolmogorov-Smirnov (KS) test \cite{berger2014kolmogorov} is a nonparametric statistical test for comparing two sample distributions to see if they come from the same overall distribution. It assesses the similarity between two samples based on the difference in the cumulative distribution function (CDF).The KS test can be used to check whether two samples follow the same distribution or whether there is a significant difference in distribution. If the two sets of data have a $\text{p-value}> \alpha$ under the KS-test, we consider that the two sets of data originate from the same distribution. Where $\alpha$ is often taken as 0.05.

\begin{figure}[t]
\centering
    \includegraphics[width = 0.8\linewidth]{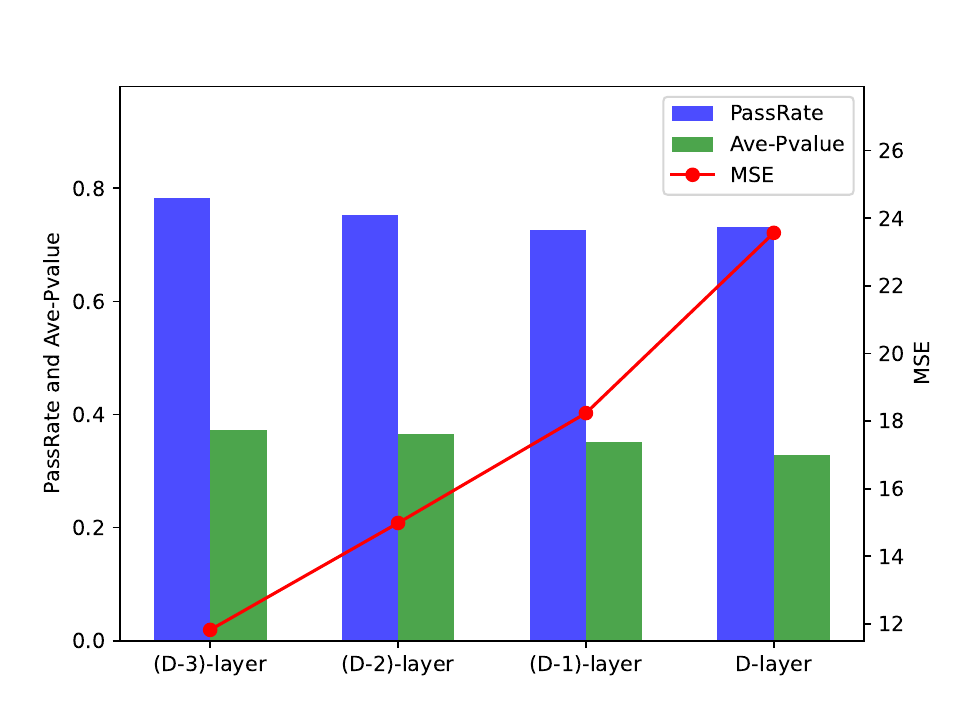}
    \caption{The first column is Original Image, another columns is Counterfactual Image generated by different Methods.}
    \label{exp2fig}
\end{figure}

From Fig. \ref{exp2fig}. It is shows experimentally that $70\%$ of the distributions fitted have a value of ks-p-value greater than or equal to 0.05, and the average ks-p-value value is 0.32. That is, in the feature space of the black-box model, we can always fit better distributions with higher confidence. And PassRate also decreases as the number of layers goes forward, i.e., the number of layers included in feature fusion is always limited. Then the rationality of our experiment is verified.

\subsection{Result and Discussion}

In Experiment 1, we juxtapose the image counterfactual explanations generated by our method with those produced by alternative approaches, both in the image space and feature space. This comparison illustrates the reasonableness and efficacy of solving for counterfactual explanations in feature space, emphasizing the importance we previously attributed to feature importance when addressing counterfactual explanations in tabular datasets. In Experiment 2, we demonstrate the viability of the core fitting distribution in our method. Intriguingly, we observe that trained black-box models consistently exhibit some form of distribution in feature space, predominantly manifesting as normal, exponential, and generalized logistic distributions.

\section{CONCLUSIONS}


For solving the counterfactual explanation of image instances, this paper proposes a novel method, DPMDCE, which proposes a new technique to compute the importance of features in the feature space of an image, on the basis of which the counterfactual explanation of features of the input image is computed in the feature space, and then the counterfactual explanation of the image is solved with the help of a generator, GAN, and a constrained optimization technique. Comparative experiments show that our method not only achieves better results on the MNIST dataset, but also greatly improves the interpretability of the image counterfactual explanation. However, our method also has limitations, as part of the optimization process is difficult to optimize for convolutional networks, so our experiments were only conducted on the single-channel dataset MNIST using fully connected neural networks. Our future work will try to address this issue to be able to make our method work on a wider range of datasets and styles of more black-box models.

\section{Acknowledgements}
This work is supported by the National Key Research and Development Program of China (Grant No.2022YFB3103702).

\bibliographystyle{splncs04}
\bibliography{reference}

\end{document}